\documentclass{article}

\usepackage{arxiv}

\usepackage[utf8]{inputenc} 
\usepackage[T1]{fontenc}    
\usepackage{hyperref}       
\usepackage{url}            
\usepackage{booktabs}       
\usepackage{amsfonts}       
\usepackage{nicefrac}       
\usepackage{microtype}      
\usepackage{lipsum}		
\usepackage{graphicx}
\usepackage{natbib}
\usepackage{doi}

\usepackage{amsmath}
\usepackage{moreverb,url}

\usepackage{algorithm}
\usepackage{algpseudocode}
\usepackage{subcaption}

\title{\texttt{ClassicLogic}: A Knowledge-Driven Benchmark of Classic Puzzle Games for Evaluating Compositional Generalization}

\author{ \href{https://orcid.org/0000-0003-0068-9998}{\includegraphics[scale=0.06]{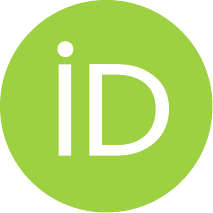}\hspace{1mm}Mahnoor Shahid}\\
	Universität Duisburg-Essen\\
	Germany \\
	\texttt{mahnoor.shahid@uni-due.de} \\
	\And
	\href{https://orcid.org/0000-0001-7727-1499}{\includegraphics[scale=0.06]{orcid.pdf}\hspace{1mm}Hannes Rothe} \\
	Universität Duisburg-Essen\\
	Germany \\
	\texttt{hannes.rothe@uni-due.de} \\
}

\date{}

\renewcommand{\headeright}{\texttt{ClassicLogic}: Benchmark of Classic Puzzle Games for Compositional Generalization}
\renewcommand{\shorttitle}{}

\hypersetup{
pdftitle={A template for the arxiv style},
pdfsubject={q-bio.NC, q-bio.QM},
pdfauthor={David S.~Hippocampus, Elias D.~Striatum},
pdfkeywords={First keyword, Second keyword, More},
}

\begin{document}
\maketitle

\begin{abstract}
Compositional generalization, the ability to understand and produce novel combinations of known components, remains a fundamental challenge for modern artificial intelligence. While few benchmarks exist, many focus on linguistic tasks and lack complex, explicit compositional structures. We introduce \textbf{\texttt{ClassicLogic}}, a new benchmark suite designed to evaluate an agent's ability to learn and compose problem-solving strategies. The benchmark consists of four classic logic puzzles: Sudoku, KenKen, Kakuro, and Futoshiki. Its core innovation is a hierarchical, explicit knowledge base for each game, where complex solving strategies are formally defined as compositions of simpler, foundational strategies. This structure allows for fine-grained evaluation of an agent's reasoning capabilities, from learning basic rules to applying multi-step compositional strategies to solve puzzles of increasing, mathematically validated difficulty. 
The open-source benchmark provides a challenging new testbed for advancing neuro-symbolic and other advanced AI reasoning systems. The benchmark is open-source and available at: \url{https://github.com/Place-Beyond-Bytes/classic_games_benchmark.git}. 
\end{abstract}


\section{Introduction}
The ability to flexibly combine existing knowledge to solve new problems is a hallmark of human intelligence \citep{sternberg1984toward,cosmides1997modular}. This capacity, termed as compositional generalization \cite{keysers2019measuring,wiedemer2023compositional}, enables us to generate a potentially infinite range of complex ideas and behaviors from a finite set of known components \citep{Fodor1988}. Modern artificial intelligence, particularly large-scale models, has demonstrated remarkable capabilities in processing and generating human-like data\citep{zhao2023survey}. Yet, a critical gap persists between their pattern-matching prowess and the robust, systematic reasoning characteristic of human intelligence\citep{marcus2018deep}. These models often fail at tasks requiring logical deduction and multi-step planning, revealing a fundamental weakness in their ability to generalize systematically \citep{patil2025advancing,cao2025large}. 

To address this, the community has developed benchmarks such as SCAN \citep{Lake2018} and COGS \citep{Kim2020} to test for compositional generalization, primarily in the domain of natural language processing. However, a gap exists for benchmarks that test compositional reasoning in more structured, symbolic problem-solving domains \citep{wang2024towards}. Such domains require not only recognizing patterns but also building and executing explicit, multi-step strategies.

In this paper, we introduce \texttt{ClassicLogic}, a novel benchmark designed to provide a transparent and challenging testbed for compositional reasoning. Our contribution is a suite of four classic logic puzzles---Sudoku, KenKen, Kakuro, and Futoshiki---that are procedurally generated and comes with a hierarchical knowledge base (KB) of solving strategies, where complex solving strategies are explicitly defined as compositions of simpler, atomic rules. For instance, advanced strategies are explicitly defined as compositions of more fundamental ones (e.g., identifying a `naked\_pair' in Sudoku is a composition of the `naked\_single' and `constraint\_propagation' strategies). This structure allows us to dissect the reasoning process and evaluate three distinct and crucial forms of compositional generalization: (1) Entity Composition, such as identifying handwritten digits as numerical entities within the puzzle grid; (2) Relational Composition, such as integrating cell-value states with arithmetic or inequality constraints; and (3) Procedural Composition, such as chaining simple solving rules into complex, multi-step strategies.

By providing a framework to test these multi-facets of reasoning, \texttt{ClassicLogic} offers a more nuanced evaluation than simple accuracy metrics. In summary, our main contributions are:

\begin{itemize}
    \item A novel, open-source benchmark (\texttt{ClassicLogic}) featuring four challenging logic puzzles with procedurally generated instances and guaranteed unique solutions.
    \item A hierarchical knowledge base (KB) for each game base, where complex strategies are explicitly composed of simpler ones, allowing for fine-grained control over task difficulty and required reasoning depth.
    \item To our knowledge, the first unified framework to explicitly disentangle and evaluate three distinct forms of compositional generalization: Entity Composition (perception), Relational Composition (rule-following), and Procedural Composition (strategic planning).
\end{itemize}

It enables researchers to diagnose why a model fails—whether it is unable to learn base rules, combine them, or transfer them. As such, \texttt{ClassicLogic} provides a crucial new tool to guide the development of more robust, interpretable, and systematically intelligent AI systems.

\section{Related Work}

The evaluation of compositional generalization has been a central theme in cognitive science and AI. Early work highlighted the systematicity of human thought as a core property of intelligence \citep{Fodor1988}, a concept that modern benchmarks aim to quantify.

\subsection{Benchmarks for Compositional Generalization}
Existing benchmarks have largely focused on linguistic and visual domains. In linguistics, datasets like SCAN \citep{Lake2018} and COGS \citep{Kim2020} evaluate whether models can generalize to novel combinations of familiar words and commands. Similarly, alongside them, others have emerged to test more complex facets. For instance, gSCAN \citep{Ruis2020} tests generalization in visually-grounded instruction following , while CFQ provides a more realistic testbed based on querying a large knowledge base \citep{Keysers2020}. In vision, the CLEVR dataset \citep{Johnson2017} evaluates for compositional reasoning about objects and their spatial and semantic relationships. Subsequent work like CLEVRER added a temporal dimension to probe causal reasoning \citep{Yi2020}, and GQA introduced a more naturalistic and compositionally complex question-answering dataset \citep{Hudson2019}. While these benchmarks are foundational for assessing generalization on perceptual and sequential data, they primarily test for an implicit understanding of composition. They are not designed to evaluate an agent's ability to learn and execute an explicit, multi-step procedural strategy from a symbolic knowledge base. However, \texttt{ClassicLogic} provides a ground-truth and symbolic hierarchy of strategies. This allows for direct evaluation of an agent's ability to build and execute complex plans in a discrete, symbolic environment. 

\subsection{AI in Games and Automated Reasoning}
The use of games as a crucible for artificial intelligence is a long and storied tradition. Landmark achievements, such as DeepMind's AlphaGo for GO \citep{Silver2016} and Pluribus for poker \citep{Brown2019}, have demonstrated superhuman performance in complex strategy games. Such immensely powerful systems often rely on reinforcement learning, trained over massive volumes of self-play and functioning as opaque, ``black-box" agents. The strategies they learn are emergent and subsymbolic, not explicitly represented in a human-interpretable format \citep{liang2025ai}. In contrast to opaque deep RL agents, \texttt{ClassicLogic} provides a transparent evaluation environment where the goal is not to create an optimal player, but to diagnose if an agent can learn and deploy human-cognizable strategies.

Moreover, there is extensive research on automated puzzle solvers, particularly for logic puzzles \citep{mitra2015learning,piette2019ludii,giadikiaroglou2024puzzle}. These problems are often formally framed as Constraint Satisfaction Problems (CSPs) \citep{Russell2010,berthier2013pattern}, for which highly efficient and specialized solvers exist. However, the goal of these systems is to find a valid solution as quickly as possible because they are optimized for the final product. They do not evaluate whether an agent can learn the human-like reasoning process required to get there \citep{qefalija2024literature}. Whereas, \texttt{ClassicLogic} is uniquely focused on evaluating the reasoning process itself. It provides a framework to test \textit{how} an agent learns, not just whether it can produce a correct answer. It is not designed to compete with bespoke CSP solvers, but rather to serve as a benchmark that tests whether a learning agent can build an understanding of the puzzle's underlying principles by learning and composing strategies from our hierarchical knowledge base. 

\subsection{Neuro-Symbolic Approaches}
A rising field that directly intersects with our work is Neuro-Symbolic (NeSy) AI. The goal of NeSy is to integrate the pattern-recognition strengths of neural networks with the reasoning capabilities of symbolic systems \citep{hitzler2022neuro,garcez2023neurosymbolic,sakr2022neuro}. Systems like DeepProblog \citep{Manhaeve2018} and Neural Logic Machines \citep{Dong2019} attempt to build models that can perform explicit logical reasoning on top of neural perception. \texttt{ClassicLogic} provides an ideal testbed for such systems, as it contains both perceptual components (e.g., recognizing MNIST digits) and a complex, symbolic reasoning structure (the strategy KB). Currently, a major challenge for the NeSy field is the lack of comprehensive benchmarks that require deep, compositional, and symbolic reasoning; we aim to fill this gap \citep{wang2024towards}. In regards to that, our benchmark serves as a testbed for NeSy systems as it provides a challenging new environment that requires the tight integration of perception and multi-step symbolic reasoning.

\section{The \texttt{ClassicLogic} Benchmark Suite}
\label{sec:benchmark}

The \texttt{ClassicLogic} benchmark is a suite of four procedurally generated logic puzzle environments designed to facilitate the diagnostic evaluation of compositional reasoning in artificial agents. In this section, we formalize its architecture, the structure of its game environments, and its core innovation: the hierarchical knowledge base of solution strategies.

\subsection{Design Principles}
The development of \texttt{ClassicLogic} is guided by four core principles to ensure a rigorous and fair evaluation platform for end-to-end reasoning systems:

\begin{itemize}
    \item \textbf{Perceptual Grounding:} All initial puzzle states are presented to the agent not as symbolic matrices, but as visual grids rendered with MNIST digit images. This design choice necessitates an initial perceptual function, $\Phi: I \rightarrow G$, to map the visual input from the image space $I$ to a symbolic grid state $G$, creating a unified testbed for integrated perception and reasoning.
    
    \item \textbf{Strategy-Driven Generation:} A puzzle instance $P$ is not generated randomly, but is constructed to require a specific, minimal set of strategies $\Sigma_{\text{req}}$ for its solution. This allows for targeted testing of an agent's ability to deploy specific reasoning patterns.
    
    \item \textbf{Validated Difficulty Scaling:} The difficulty of a puzzle $P$, denoted by a function $D(P)$, is directly correlated with the compositional depth of the strategies in $\Sigma_{\text{req}}$. We define compositional depth, $\delta(s)$, of a strategy $s$ recursively, providing a formal measure of puzzle complexity.
    
    \item \textbf{Guaranteed Uniqueness:} For every puzzle instance $P$ generated by the benchmark, there exists a unique solution state $S_{\text{sol}}$. This eliminates ambiguity in evaluation and ensures that success is based on valid logical deduction.
\end{itemize}

\subsection{Game Environments}
The suite comprises four classic logic puzzles, each formalized as a distinct environment, as mentioned in Table \ref{tab:game_overview}. Let a grid state be a matrix $G$ of size $N \times N$.
The state of any puzzle, $P$, can be represented as a tensor $T \in \mathbb{R}^{N \times N \times K}$, where $N$ is the grid dimension and $K$ represents the features of each cell (e.g., a one-hot vector for candidate digits). The action space $\mathcal{A}$ is typically discrete, allowing an agent to set a value $v$ in a cell $(i, j)$, i.e., $\mathcal{A} \subseteq \{ (i, j, v) \}_{i,j,v=1}^{N}$. 

\begin{table}[h!]
\small\sf\centering
\caption{Overview of the Game Environments in the \texttt{ClassicLogic} Benchmark Suite.\label{tab:game_overview}}
\begin{tabular}{llll}
\toprule
\textbf{Game} & \textbf{Grid Size} & \textbf{Core Constraint Type} \\
\midrule
\texttt{Sudoku} & $9 \times 9$ & Latin Square + Block  \\
\texttt{KenKen} & $N \times N$ & Latin Square + Arithmetic Cages  \\
\texttt{Kakuro} & Irregular & Cross-Sum Logic \\
\texttt{Futoshiki} & $N \times N$ & Latin Square + Inequality  \\
\bottomrule
\end{tabular}
\end{table}

\paragraph{Sudoku}
The environment is a partially filled $9 \times 9$ grid $G$. The objective is to fill the empty cells with digits $d \in \{1, ..., 9\}$ subject to the constraint that for any cell $G_{ij}$, its value must be unique in its row, column, and designated $3 \times 3$ subgrid. Formally, $\forall i, j, k \in \{1, ..., 9\}$ where $i \neq k$, $G_{ij} \neq G_{kj}$; where $j \neq k$, $G_{ij} \neq G_{ik}$; and for all cells $(i', j')$ in the same $3 \times 3$ block as $(i, j)$, $G_{ij} \neq G_{i'j'}$.

\paragraph{KenKen}
An $N \times N$ grid is partitioned into a set of non-overlapping cages $C = \{c_1, c_2, ..., c_m\}$. Each cage $c_k \in C$ is defined by a tuple $(\mathcal{Z}_k, V_k, O_k)$, where $\mathcal{Z}_k$ is the set of cell coordinates in the cage, $V_k \in \mathbb{N}^+$ is a target value, and $O_k \in \{+, -, \times, \div\}$ is a binary arithmetic operator. The constraint for each cage is that the application of operator $O_k$ to the values in the cells of $\mathcal{Z}_k$ must result in $V_k$.

\paragraph{Kakuro}
The grid consists of ``clue cells" and ``entry cells." A clue cell contains one or two clue values, $(c_{\text{down}}, c_{\text{right}})$, where $c \in \mathbb{N}^+$. The constraint is that the sum of the digits in the contiguous block of entry cells below the clue must equal $c_{\text{down}}$, and the sum of the digits in the block to its right must equal $c_{\text{right}}$. Furthermore, all digits within any given block must be unique.

\paragraph{Futoshiki}
An $N \times N$ grid must be filled with digits $\{1, ..., N\}$ adhering to Latin square rules (uniqueness in rows and columns). Additionally, the environment contains a set of relational constraints $\mathcal{R}$ of the form $G_{ij} \prec G_{i'j'}$, where $(i,j)$ and $(i',j')$ are adjacent cell coordinates and $\prec \in \{<, >\}$.

\begin{figure}[t]
    \centering
    \begin{subfigure}[b]{0.3\columnwidth}
        \centering
        \includegraphics[width=\linewidth]{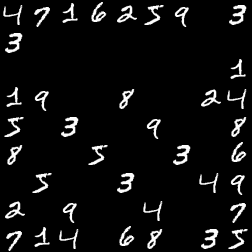}
        \caption{Sudoku Puzzle}
    \end{subfigure}
    \hfill
    \begin{subfigure}[b]{0.3\columnwidth}
        \centering
        \includegraphics[width=\linewidth]{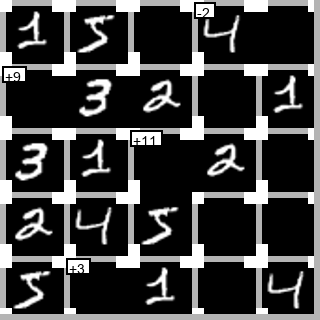}
        \caption{KenKen Puzzle}
    \end{subfigure}
    \hfill
    \begin{subfigure}[b]{0.3\columnwidth}
        \centering
        \includegraphics[width=\linewidth]{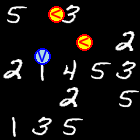}
        \caption{Futoshiki Puzzle}
    \end{subfigure}
    
    \vspace{0.2cm} 
    
    \begin{subfigure}[b]{0.3\columnwidth}
        \centering
        \includegraphics[width=\linewidth]{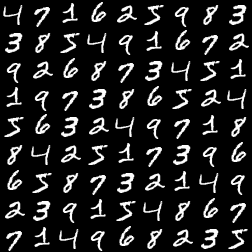}
        \caption{Sudoku Solution}
    \end{subfigure}
    \hfill
    \begin{subfigure}[b]{0.3\columnwidth}
        \centering
        \includegraphics[width=\linewidth]{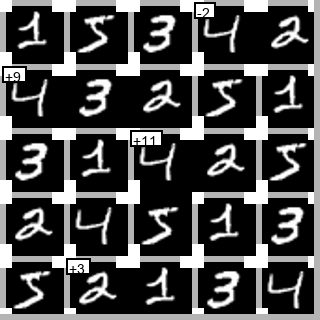}
        \caption{KenKen Solution}
    \end{subfigure}
    \hfill
    \begin{subfigure}[b]{0.3\columnwidth}
        \centering
        \includegraphics[width=\linewidth]{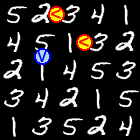}
        \caption{Futoshiki Solution}
    \end{subfigure}
    
    \caption{Visual examples from the \texttt{ClassicLogic} benchmark, showcasing the diversity of reasoning challenges. The top row (a-c) displays the initial puzzle states for Sudoku, KenKen, and Futoshiki, presented with perceptual MNIST digits. The bottom row (d-f) shows their corresponding, unique logical solutions.}
    \label{fig:puzzle_samples}
\end{figure}

To provide a concrete visualization of these environments, Figure \ref{fig:puzzle_samples} presents sample instances for Sudoku, KenKen, and Futoshiki. Each example displays both the initial puzzle state, as it would be presented to an agent, and its corresponding unique solution. These samples illustrate the diversity of constraints and logical structures that agents must handle within the \texttt{ClassicLogic} benchmark.

\subsection{The Hierarchical Knowledge Base (KB)}
The central innovation of \texttt{ClassicLogic} is its explicit, hierarchical knowledge base. Formally, the KB for each game is a tuple $\mathcal{K} = (\mathcal{S}, \mathcal{C})$, where $\mathcal{S}$ is a set of strategies and $\mathcal{C}$ is a composition relation defined on $\mathcal{S}$.

The set of strategies $\mathcal{S}$ is partitioned into a set of atomic base strategies $\mathcal{S}_{\text{base}}$ and a set of composed strategies $\mathcal{S}_{\text{comp}}$. A strategy $s \in \mathcal{S}$ is a function that maps a puzzle state $P_t$ to a new, more constrained state $P_{t+1}$ by making a valid logical deduction, i.e., $s: P_t \mapsto P_{t+1}$.

The composition relation $\mathcal{C}$ defines how strategies in $\mathcal{S}_{\text{comp}}$ are constructed. For any composed strategy $s_c \in \mathcal{S}_{\text{comp}}$, there exists an ordered set of constituent strategies $\{s_1, s_2, ..., s_k\} \subseteq \mathcal{S}$ such that $s_c$ represents the sequential application or logical combination of these constituents. We can denote this composition as $s_c := \bigoplus_{i=1}^{k} s_i$.

\begin{figure}[t]
	\centering
	\includegraphics[width=0.9\linewidth]{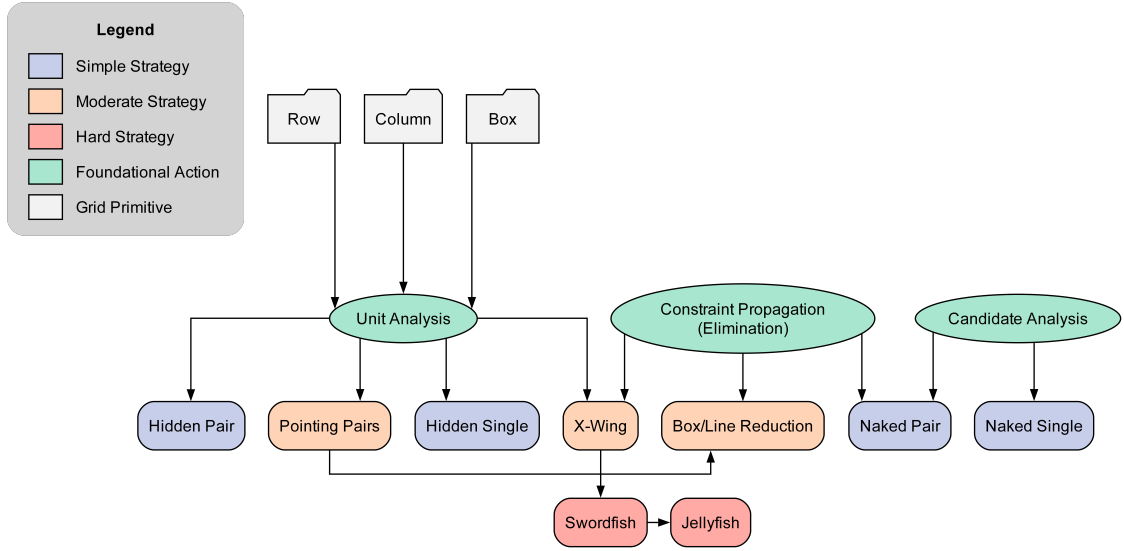} 
	\caption{An example illustration of the strategy hierarchy in the \texttt{ClassicLogic} Knowledge Base of Sudoku. Complex strategies are explicitly constructed from simpler, foundational ones, allowing for precise control over the compositional reasoning required by a puzzle.}
	\label{fig:kb_hierarchy}
\end{figure}

For example, consider the state of a Sudoku puzzle $P$ as a grid of candidate sets, where $P_{ij} \subseteq \{1, ..., 9\}$ is the set of possible values for cell $(i, j)$. A base strategy, \textit{Constraint Propagation} ($s_{\text{cp}}$), takes a confirmed cell value and removes it from the candidate sets of all peer cells. This simple action is a foundational component for more complex strategies. A moderate strategy like \textit{Naked Pair} ($s_{\text{np}}$) is composed of a pattern-recognition step followed by this action, which we can denote as $s_{\text{np}} := s_{\text{find\_pair}} \oplus s_{\text{cp}}$. The sub-procedure, $s_{\text{find\_pair}}$, identifies when two cells in a unit share the exact same two candidates (e.g., $P_{ij} = P_{ik} = \{a, b\}$). This finding then enables the application of $s_{\text{cp}}$ to eliminate candidates $a$ and $b$ from all other cells in that unit.

This compositional principle extends to more advanced, multi-unit strategies as visualized in Figure \ref{fig:kb_hierarchy}. An \textit{X-Wing} ($s_{\text{xw}}$) is a more complex moderate strategy that is similarly composed of a sophisticated pattern-recognition procedure followed by the same base action: $s_{\text{xw}} := s_{\text{find\_xwing}} \oplus s_{\text{cp}}$. Here, $s_{\text{find\_xwing}}$ is itself a composition of a more basic procedure, let's call it \textit{Unit Scan} ($s_{\text{scan}}$), which finds the positions of a candidate in a unit. An X-Wing pattern is identified for a digit $d$ when there exist two rows $r_1, r_2$ and two columns $c_1, c_2$ such that the repeated application of $s_{\text{scan}}$ confirms that $\text{Positions}(d, r_1) = \{c_1, c_2\}$ and $\text{Positions}(d, r_2) = \{c_1, c_2\}$.

The hierarchy deepens further with hard strategies like \textit{Swordfish} ($s_{\text{sf}}$), which is a direct extension of the X-Wing logic. It is composed as $s_{\text{sf}} := s_{\text{find\_swordfish}} \oplus s_{\text{cp}}$, where $s_{\text{find\_swordfish}}$ is a scaled-up version of the X-Wing's pattern recognition. It applies the same foundational $s_{\text{scan}}$ procedure but seeks a 3x3 pattern across three rows and three columns. This demonstrates a deep compositional structure where advanced strategies are formed not just by combining different simple rules, but by scaling and reapplying the same underlying logical patterns in a more complex configuration.

\subsection{Procedural Generation and Difficulty Calibration}
\label{sec:generation}

A central requirement for a robust benchmark is the ability to generate a vast and diverse set of puzzles with precisely controlled properties. Standard reverse-solving algorithms, which iteratively remove clues from a solved grid, are often computationally intractable. To address this, we employ a sophisticated two-stage hybrid generation process. This methodology separates the computationally expensive task of logical structure generation from the lightweight task of producing playable instances, providing both rigor and real-time performance.

\subsubsection{Stage 1: Strategy-Driven Template Generation}
The goal of this stage is to produce a library, $\mathbb{TPL}$, of minimal, abstract puzzle structures, or templates. Each template represents the logical essence of a puzzle class with a specific difficulty.

We formally define a puzzle instance as a tuple $P = (\mathcal{G}, \mathcal{C}_P)$, where $\mathcal{G}$ is a grid structure and $\mathcal{C}_P$ is a set of concrete clues (e.g., pre-filled cell values, arithmetic cage definitions). A solution $S$ is a complete assignment of values to $\mathcal{G}$ that satisfies all game rules and the clues in $\mathcal{C}_P$.

A template is an abstracted puzzle, $Tpl = (\mathcal{L}, \mathcal{C}_{\text{abs}}, \Sigma_{\text{req}})$, where $\mathcal{L}$ is a logical layout (e.g., a graph representation of the grid), $\mathcal{C}_{\text{abs}}$ is a set of abstract placeholder constraints, and $\Sigma_{\text{req}}$ is the pre-analyzed, minimal set of strategies required to solve any valid instantiation of this template.

Template generation (Algorithm \ref{alg:template_gen}) begins with a canonical solution grid, $S_{\text{sol}}$. It then performs an iterative clue ablation process. For each clue $c \in \mathcal{C}_{S_{\text{sol}}}$, it creates a candidate puzzle $P' = (\mathcal{G}, \mathcal{C}_{P} \setminus \{c\})$. This candidate is validated against two criteria using two distinct solvers:
\begin{enumerate}
    \item \textbf{Uniqueness Check:} A complete, optimal solver, $\texttt{FullSolver}(P')$, which returns the set of all valid solutions. The removal is valid only if $| \texttt{FullSolver}(P') | = 1$.
    \item \textbf{Difficulty Check:} A constrained solver, $\texttt{LimitedSolver}(P', \mathcal{K}_{\leq \mathcal{D}_{\text{target}}})$, equipped only with strategies from the Knowledge Base whose compositional depth $\delta(s)$ is less than or equal to the target difficulty $\mathcal{D}_{\text{target}}$. The removal is valid only if this solver finds the solution, ensuring the puzzle does not require harder strategies.
\end{enumerate}

The final, minimal puzzle $P_{\text{final}}$ is then abstracted into a canonical template $Tpl$ by a function $f_{\text{abs}}: P \rightarrow Tpl$. This intensive process is run to populate the template library $\mathbb{TPL}$. 
To validate our formal difficulty metric, we conducted an empirical study correlating $D(P)$ with the performance of our baseline heuristic solver (see Section \ref{sec:validation}).

\begin{algorithm}[h!]
\caption{Strategy-Driven Template Generation}
\label{alg:template_gen}
\begin{algorithmic}[1]
\State \textbf{Input:} Target difficulty $\mathcal{D}_{\text{target}}$
\State \textbf{Initialize:} $S_{\text{sol}} \gets \texttt{GenerateCanonicalSolution}()$
\State $P \gets (\mathcal{G}, \mathcal{C}_{S_{\text{sol}}})$ \Comment{Initial puzzle is the full solution}
\State $\mathcal{C}_{\text{removable}} \gets \texttt{GetClues}(P)$

\For{each clue $c \in \texttt{Shuffled}(\mathcal{C}_{\text{removable}})$}
    \State $P' \gets (\mathcal{G}, \mathcal{C}_{P} \setminus \{c\})$
    \If{$| \texttt{FullSolver}(P') | = 1$}
        \State Let $\mathcal{K}_{\text{limit}} \gets \{s \in \mathcal{K} \mid \delta(s) \leq \mathcal{D}_{\text{target}}\}$
        \If{$\texttt{LimitedSolver}(P', \mathcal{K}_{\text{limit}})$ is successful}
            \State $P \gets P'$ \Comment{Accept clue ablation}
        \EndIf
    \EndIf
\EndFor
\State $Tpl \gets f_{\text{abs}}(P)$ \Comment{Abstract final puzzle to a template}
\State \textbf{Output:} Template $Tpl$
\end{algorithmic}
\end{algorithm}

\subsubsection{Stage 2: Real-time Puzzle Instantiation}
With a rich library $\mathbb{TPL}$ of pre-generated and validated templates, creating a playable puzzle instance is computationally trivial. This process (Algorithm \ref{alg:puzzle_instantiation}) involves template selection followed by instantiation and transformation.

The instantiation function, $f_{\text{inst}}: Tpl \rightarrow P_{\text{base}}$, maps the template's abstract structure $\mathcal{L}$ and constraints $\mathcal{C}_{\text{abs}}$ to a concrete puzzle instance $P_{\text{base}}$. This is achieved via a randomized backtracking search that assigns concrete values while satisfying the template's constraints.

To generate a wide variety of structurally equivalent but superficially distinct puzzles from a single template, we apply a transformation $\tau$ from a set of symmetry-preserving transformations $\mathcal{T}$ (e.g., rotation, reflection, and value permutations). The final puzzle is $P_{\text{final}} = \tau(P_{\text{base}})$. This two-stage architecture ensures both logical rigor and real-time performance.

\begin{algorithm}[h!]
\caption{Real-time Puzzle Instantiation}
\label{alg:puzzle_instantiation}
\begin{algorithmic}[1]
\State \textbf{Input:} Target difficulty $\mathcal{D}_{\text{target}}$, Template library $\mathbb{TPL}$
\State \textbf{Initialize:} $\mathbb{TPL}_{\text{filtered}} \gets \{Tpl \in \mathbb{TPL} | D(Tpl) = \mathcal{D}_{\text{target}}\}$

\If{$\mathbb{TPL}_{\text{filtered}}$ is empty}
    \State \textbf{return} \texttt{Error("No templates for difficulty")}
\EndIf

\State $Tpl_{\text{selected}} \gets \texttt{RandomlySelect}(\mathbb{TPL}_{\text{filtered}})$
\State $P_{\text{base}} \gets f_{\text{inst}}(Tpl_{\text{selected}})$
\State $\tau \gets \texttt{RandomlySelect}(\mathcal{T})$ \Comment{Select a transformation}
\State $P_{\text{final}} \gets \tau(P_{\text{base}})$ \Comment{Apply transformation}
\State \textbf{Output:} Final puzzle instance $P_{\text{final}}$
\end{algorithmic}
\end{algorithm}

\subsection{A Framework for Evaluating Compositional Generalization}
\label{sec:eval_compositionality}

The \texttt{ClassicLogic} benchmark provides a structured framework to evaluate three distinct and critical forms of compositional generalization. We formalize these as follows, from basic perception to multi-step strategic planning.

\paragraph{1. Entity Composition}
is the foundational ability to learn a mapping function, $\Phi: I \rightarrow E$, from the raw perceptual input space $I$ (pixel grids) to a set of symbolic entities $E$. An entity $e \in E$ is a tuple representing a fundamental object, such as $(\text{digit}, 7)$ or $(\text{operator}, >)$. An agent demonstrates successful entity composition if, for every initial clue image $i_{r,c}$ at grid location $(r,c)$, its inferred entity $\hat{e}_{r,c} = \Phi(i_{r,c})$ equals the ground-truth entity $e_{r,c}$.

\begin{itemize}
    \item \textbf{Example 1 (KenKen):} An agent is presented with a visual cage containing the clue ``7+". Successful composition requires generating the entity set $\{(\text{digit}, 7), (\text{operator}, +)\}$. An output of $\{(\text{digit}, 1), (\text{operator}, +)\}$ would represent a critical failure in entity recognition.
    
    \item \textbf{Example 2 (Futoshiki):} Given a grid of MNIST digits and $>$ symbols, a successful agent must parse the entire initial state $P_0$ into a set of symbolic facts, such as $\{\text{Value}(r_1, c_1, 5), \text{Constraint}(r_2, c_2, >) ...\}$. Misclassifying a handwritten digit or an operator indicates a failure at this basic compositional level.
\end{itemize}

\paragraph{2. Relational Composition}
is the ability to apply a set of universal relational rules, $\mathcal{R}$, to the set of entities $E$ to infer new facts or validate state transitions. A rule $R \in \mathcal{R}$ is a predicate over one or more entities that must hold true for a state to be valid. An agent's action $a_t$ at state $S_t$ is valid only if the resulting state $S_{t+1}$ satisfies all rules, i.e., $\forall R \in \mathcal{R}, R(S_{t+1}) = \text{true}$. An illegal move signals a failure in relational composition.

\begin{itemize}
    \item \textbf{Example 1 (Futoshiki):} Given the symbolic facts $\text{Constraint}(c_1, c_2, >)$ and $\text{Value}(c_2, 4)$, the agent must apply the rule: 
    $$ \forall c_1, c_2, v_1, v_2: (\text{Value}(c_1, v_1) \land \text{Value}(c_2, v_2) \land \text{Constraint}(c_1, c_2, >)) \rightarrow v_1 > v_2 $$
    By composing the rule with the known facts, a successful agent infers that the candidate set for $c_1$ is now constrained, $\text{Candidates}(c_1) \subseteq \{5,6,7,8,9\}$.
    
    \item \textbf{Example 2 (Sudoku):} An agent in a state with $\text{Value}(1,2, 7)$ considers the action $a_t = \text{place}(1,5,7)$. To validate this, it must apply the row-uniqueness rule:
    $$ \forall r, c_1, c_2, v: (c_1 \neq c_2 \land \text{Value}(r, c_1, v)) \rightarrow \neg \text{Candidate}(v, r, c_2) $$
    Since this rule is violated by the proposed action, a successful agent would prune this action from its policy.
\end{itemize}

\paragraph{3. Procedural Composition}
This is the highest form of compositionality, representing the ability to construct a new procedure (strategy), $s_{\text{new}}$, by sequencing a set of known, simpler procedures $\{s_1, ..., s_k\} \subseteq \mathcal{S}_{\text{known}}$. Formally, $s_{\text{new}} := s_k \circ ... \circ s_1$, where $\circ$ represents sequential application or functional composition. We train an agent on puzzles $\mathcal{P}_{\text{base}}$ solvable only by procedures in $\mathcal{S}_{\text{base}}$. We then test it on puzzles $\mathcal{P}_{\text{comp}}$ that require at least one procedure $s_c \notin \mathcal{S}_{\text{base}}$, where $s_c$ is a known composition of procedures from $\mathcal{S}_{\text{base}}$. Success on $\mathcal{P}_{\text{comp}}$ provides direct evidence of procedural composition.

\begin{itemize}
    \item \textbf{Example 1 (Sudoku):} An agent knows the procedure $s_{\text{pattern\_rec}}$ (recognize when two cells in a row have candidates $\{2,8\}$) and $s_{\text{elimination}}$ (apply constraint propagation). It must compose these to form a novel strategy, $s_{\text{naked\_pair}} := s_{\text{elimination}} \circ s_{\text{pattern\_rec}}$, where the output of the pattern recognition (the cells and candidates to eliminate) becomes the input for the elimination procedure.
    
    \item \textbf{Example 2 (Kakuro):} An agent knows $s_{\text{unique\_sum}}$ (e.g., a 2-cell sum of 3 must be $\{1,2\}$) and $s_{\text{cross\_ref}}$ (use a confirmed value to constrain an intersecting block). Faced with intersecting blocks, it must compose a new procedure: first, apply $s_{\text{unique\_sum}}$ to reduce one block's candidates to $\{1,2\}$; then, apply $s_{\text{cross\_ref}}$ using a known digit from the second block to disambiguate the first. This chain of dependent procedures is a hallmark of procedural composition.
\end{itemize}

\subsection{Implementation and Accessibility}
To maximize reproducibility and ease of adoption, the \texttt{ClassicLogic} benchmark is implemented with a strong focus on accessibility and a streamlined setup process. The entire suite is developed in Python 3 and its dependencies are explicitly managed. A key feature is the unified environment setup. We provide automated scripts designed to create a self-contained virtual environment and install all necessary packages with a single command. The repository includes `\texttt{unified\_env\_setup.py}' (for POSIX systems) and \texttt{unified\_env\_quick\_setup.bat} (for Windows), which handle the environment creation and dependency installation listed in the `requirements.txt' file. This eliminates common setup hurdles and ensures that researchers can begin experimentation with minimal friction. Furthermore, environment configurations are managed through a central `classic\_games\_env\_info.json' file. This file contains metadata and settings for the benchmark, allowing for easy modification and inspection of the environment's parameters without altering the core source code. The framework is also designed with modularity and extensibility in mind. Adding a new puzzle to the suite follows a well-defined template, primarily requiring the implementation of the game's state transition logic and its corresponding strategy knowledge base. This structure encourages community contributions and the long-term expansion of \texttt{ClassicLogic} with new reasoning challenges.

The entire framework, including the source code, setup scripts, and documentation, is publicly available under an MIT License on GitHub\footnote{\url{https://github.com/Place-Beyond-Bytes/classic_games_benchmark.git}} to encourage open collaboration and extension for the research community.

\section{Experiments}
\label{sec:experiments}

We conducted a set of experiments to demonstrate how this benchmark can be utilized to evalute an agent's compositional reasoning abilities using \texttt{ClassicLogic}. 


\subsection{Preliminary Validation of the Difficulty Metric}
\label{sec:validation}

\paragraph{Objective}
Before using \texttt{ClassicLogic} to evaluate AI agents, we first conducted a preliminary experiment to empirically validate that our theoretical difficulty metric, $D(P)$, derived from the Knowledge Base, correlates with practical computational hardness. To establish both internal and external validity, we measured performance using two distinct automated solvers.

\paragraph{Methodology}
For this validation study, we focused on Sudoku as a representative environment, given its well-understood properties and rich strategy space. We generated a validation set of 99 Sudoku puzzles, comprising 33 unique instances for each difficulty category (Easy, Moderate, Hard). We then used two distinct automated solvers to solve each puzzle:

\begin{itemize}
    \item \textbf{Glass-Box Solver:} Our implemented baseline agent that greedily applies strategies from the Knowledge Base $\mathcal{K}$. We measured the total number of distinct strategies applied to reach a solution.
    \item \textbf{Black-Box Solver:} A standard, optimized Constraint Satisfaction Problem (CSP) solver. This solver is unaware of our KB. We measured the wall-clock solve time in milliseconds.
\end{itemize}

Our hypothesis is that as the difficulty category progresses from Easy to Hard, the metrics from both solvers will increase significantly. We treated the categories as ranks (Easy=1, Moderate=2, Hard=3) and computed the Spearman's rank correlation coefficient ($\rho$).

\begin{figure}[h!]
    \centering
    \includegraphics[width=0.75\linewidth]{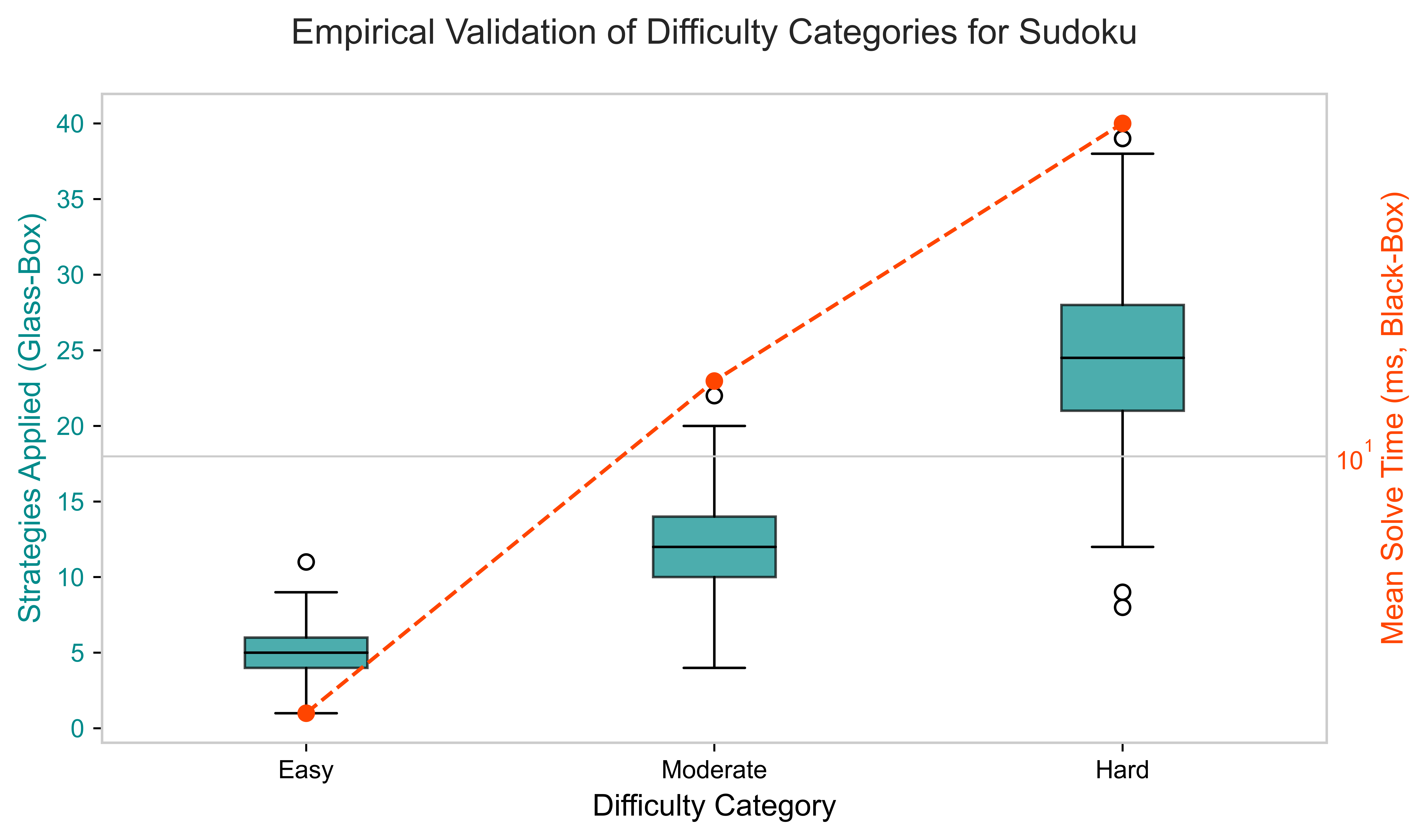}
    \caption{Correlation between the difficulty categories and empirical measures. The bars (left axis) show the increase in strategies applied by our ``glass-box" solver, while the line plot (right axis) shows the exponential increase in solve time for an independent ``black-box" CSP solver.}
    \label{fig:difficulty_validation}
\end{figure}

\paragraph{Results}
The results, summarized in Table \ref{tab:difficulty_validation} and visualized in Figure \ref{fig:difficulty_validation}, confirm our hypothesis. The number of strategies applied by the glass-box solver and the solve time for the black-box CSP solver both increase substantially with each difficulty category, demonstrating a strong, positive, and monotonic relationship.

\begin{table}[h!]
\centering
\caption{Empirical validation results for the three difficulty categories using the Sudoku environment. Mean strategies applied ("glass-box" solver) and mean solve time ("black-box" solver) are shown. Standard deviations are in parentheses.}
\label{tab:difficulty_validation}
\begin{tabular}{l c c}
\toprule
\textbf{Difficulty Category} & \textbf{Mean Strategies Applied} & \textbf{Mean Solve Time (ms)} \\
\midrule
\textbf{Easy} & 5.4 (\textpm 1.5) & 2.1 (\textpm 0.6) \\
\textbf{Moderate} & 12.8 (\textpm 3.2) & 15.8 (\textpm 4.2) \\
\textbf{Hard} & 24.7 (\textpm 5.9) & 75.6 (\textpm 18.9) \\
\bottomrule
\end{tabular}
\end{table}

The strong correlation with an independent, black-box solver ($\rho = 0.98, p < 0.01$) provides the evidence that our formal difficulty metric is a valid proxy for intrinsic computational hardness. This validation of our methodology on a representative puzzle provides confidence that the difficulty scaling is a robust feature of the entire benchmark suite.

\subsection{Evaluating Compositional Generalization}
We evaluated three representative neuro-symbolic (NeSy) models using the \texttt{ClassicLogic} benchmark to assess their compositional generalization capabilities. The models were chosen to represent different approaches to integrating neural and symbolic reasoning:

\begin{itemize}
    \item \textbf{SATNet} \cite{Wang2019}: An architecture that integrates a differentiable MAX-SAT solver, allowing a neural network to learn the logical constraints of a problem.
    \item \textbf{Neural Module Network (NMN)} \cite{Andreas2016}: A model that dynamically assembles a custom neural network from smaller, specialized modules to create a task-specific reasoning pipeline.
    \item \textbf{Logical Neural Network (LNN)} \cite{Riegel2020}: A framework where each neuron has a clear logical meaning (e.g., AND, OR), creating a highly interpretable network that adheres to formal logic.
\end{itemize}

The following experiments are intended to demonstrate the benchmark's utility in evaluating and differentiating the capabilities of various models across all four game environments. We conducted two distinct sets of experiments to probe different facets of compositional generalization. First, in Section 4.3, we evaluated the ability of models to perform zero-shot transfer of perceptual and relational knowledge from a single training game to three unseen games. Second, in Section 4.4, we test for in-domain procedural generalization, evaluating if models can compose known strategies into novel, more complex ones within each game environment.

\subsection{Entity and Relational Composition Performance}

\paragraph{Training Protocol}
To rigorously test for zero-shot generalization, all models were trained exclusively on the Sudoku environment. For Entity Composition, the models' perceptual frontends were trained to recognize the MNIST digits used as clues in Sudoku puzzles. For Relational Composition, the models were trained only on Sudoku's specific rule set: the uniqueness constraint within rows, columns, and 3x3 blocks. Their performance was then evaluated in a zero-shot setting on the unseen visual and logical structures of KenKen, Kakuro, and Futoshiki.

\paragraph{Results}
The results, summarized in Table \ref{tab:entity_relational_results}, reveal the profound challenge of out-of-domain generalization for these models. In Entity Composition, while all models learned to parse Sudoku digits well, their accuracy dropped significantly on the other games. The LNN performed best, maintaining accuracy in the low 80s, but all models struggled, indicating that their visual representations were not abstract enough to generalize to the novel layouts and symbols (e.g., cages, operators) in the other puzzles.
The performance on Relational Composition tells a similar story. The models found it extremely difficult to transfer the learned Sudoku rules to the fundamentally different constraints of the other games. The LNN was the most robust, maintaining the lowest illegal move rate, which is consistent with its logic-based architecture. However, even it made frequent errors when faced with the arithmetic and ordinal logic of KenKen and Futoshiki. SATNet and NMN struggled even more, with illegal move rates reaching as high as 45.5\% for the NMN in the Kakuro environment. This demonstrates a critical failure to generalize abstract logical principles, a key weakness that this benchmark effectively exposes.

\begin{table*}[b]
\centering
\caption{Entity and Relational composition results across all four game environments. Models were trained only on Sudoku. The results show that models are generalizing well across other games.}
\label{tab:entity_relational_results}
\begin{tabular}{l c c c c c c c c}
\toprule
& \multicolumn{4}{c}{\textbf{Entity Parsing Accuracy (\%)}} & \multicolumn{4}{c}{\textbf{Illegal Move Rate (\%)}} \\
\cmidrule(lr){2-5} \cmidrule(lr){6-9}
\textbf{Model} & \textbf{Sudoku} & \textbf{KenKen} & \textbf{Kakuro} & \textbf{Futoshiki} & \textbf{Sudoku} & \textbf{KenKen} & \textbf{Kakuro} & \textbf{Futoshiki} \\
\midrule
SATNet & 79.7 & 70.6 & 74.7 & 70.5 & 6.8 & 17.8 & 25.0 & 9.5 \\
NMN & 67.5 & 60.4 & 61.7 & 61.4 & 15.7 & 31.8 & 45.5 & 17.1 \\
LNN & \textbf{86.8} & \textbf{82.7} & \textbf{80.8} & \textbf{84.7} & \textbf{5.8} & \textbf{6.8} & \textbf{10.0} & \textbf{5.5} \\
\bottomrule
\end{tabular}
\end{table*}

\subsection{Procedural Composition Performance}

\paragraph{Training Protocol}
This experiment tests the most challenging facet of reasoning: composing known rules into novel, multi-step strategies. Unlike the previous zero-shot transfer task, this is a test of in-domain procedural generalization. For each of the four games, all three models were trained on a curriculum of ``Easy" and ``Moderate" puzzles. This training corpus exposed them to all the base and moderate-level strategies in the Knowledge Base. The models were then tested on a set of ``Hard" puzzles, which are guaranteed to require composed strategies in sequences not seen during training.

\begin{figure*}[b]
    \centering
    \includegraphics[width=0.78\linewidth]{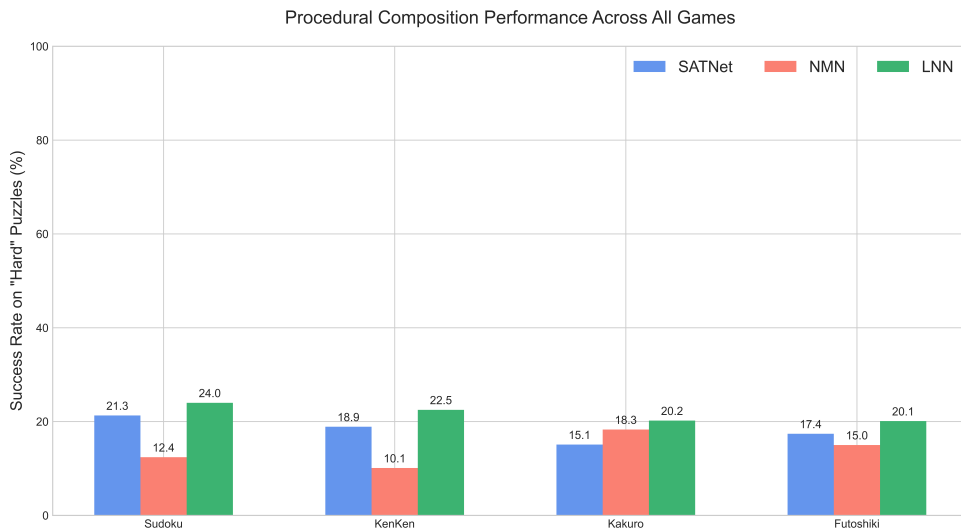}
    \caption{Procedural composition results across all games. Success rate defines the accurate formation and application of ``Hard" strategies required to solve the game. A dramatic "generalization gap" can be observed as a key failure in procedural composition.}
    \label{fig:procedural_results}
\end{figure*}

\paragraph{Results}
The results, shown in Figure \ref{fig:procedural_results}, are striking and reveal a critical failure point for all tested architectures. Despite their good performance on adhering to static rules (Relational Composition), all models showed an inability to compose those rules into novel, multi-step strategies. The success rates on the Hard puzzles collapsed to below 25\% across the board.

This demonstrates a stark disconnect between knowing the rules of a game and knowing how to play it strategically. The LNN, which was the most robust in the relational task, remains the top performer here, but its low success rate (20-24\%) underscores the immense difficulty of the challenge. The NMN, which might have been expected to perform well due to its modularity, struggled the most, suggesting its dynamic assembly process failed to generalize to more complex procedures. These results powerfully illustrate that even models with strong logical priors are not inherently equipped for the flexible, sequential planning required by our benchmark. 

\section{Discussion}
Our experiments demonstrate that \texttt{ClassicLogic} can effectively diagnose a critical failure mode in modern neuro-symbolic systems. The ability to pinpoint this specific gap between relational and procedural generalization is a direct result of the benchmark's design, which provides a unified framework for evaluating these distinct layers of compositionality in a single, coherent environment. While models adept at learning static facts and rules, they remain brittle at composing those rules into dynamic, multi-step procedures. The severe performance collapse on ``Hard" puzzles is not merely a quantitative failure; it is a qualitative one that exposes a deep gap between pattern recognition and true strategic reasoning.

This finding suggests that many current architectures, even those with strong logical priors like LNNs and SATNet, lack the mechanisms for flexible, sequential planning. Knowing the rules of chess (relational composition) is fundamentally different from knowing how to play it well (procedural composition). 

Ultimately, this validates \texttt{ClassicLogic} not just as another benchmark, but as a necessary course correction for the field. It challenges the community to move beyond tasks that can be solved with static rule application and to focus on the more difficult, more human-like skill of composing knowledge into novel plans. We believe that progress on the kind of challenges presented in our benchmark is a prerequisite for building the next generation of robust, trustworthy, and truly intelligent AI systems.

\section{Limitations}
While \texttt{ClassicLogic} provides a comprehensive framework, it is important to acknowledge its scope and inherent limitations to properly situate its contributions and guide future work. The four puzzles in the suite were selected to cover a diverse range of logical constraints (e.g., arithmetic, relational, spatial). However, they represent only a subset of the vast world of logic puzzles. The benchmark does not currently include other families such as graph-coloring puzzles, temporal logic puzzles, or those requiring extensive common-sense knowledge. Moreover, we have curated a hierarchically structured KB for each game based on well-documented human strategies. However, we do not claim this KB is exhaustive. Humans may employ novel or idiosyncratic strategies not captured in our formal system. 

These limitations do not diminish the benchmark's utility but rather clarify its intended purpose: to provide a deep, diagnostic tool for a specific, vital, and currently under-evaluated facet of artificial intelligence. We believe these focused evaluations are a prerequisite for building more general and robust AI systems, and we welcome community contributions to expand the benchmark's scope in the future.

\section{Conclusion}

In this work, we introduced \texttt{ClassicLogic}, which to best of our knowledge is the first benchmark to provide a unified framework for evaluating the full stack of compositional reasoning: from Entity Composition (perception) to Relational Composition (rule-following) and finally to Procedural Composition (strategic planning). Our illustrative experiments revealed a profound gap: while current neuro-symbolic models excel at adhering to static rules, they fail catastrophically when required to compose known procedures into novel, multi-step plans. This finding validates \texttt{ClassicLogic} as a crucial diagnostic tool that moves beyond simple rule-following to probe the deeper challenge of strategic reasoning. By providing a clear and challenging measure of procedural generalization, we hope to guide the community toward building the next generation of AI that can not only learn facts but can compose them into robust and trustworthy strategies.



\bibliographystyle{apalike}
\bibliography{references}

@article{sternberg1984toward,
  title={Toward a triarchic theory of human intelligence},
  author={Sternberg, Robert J},
  journal={Behavioral and Brain Sciences},
  volume={7},
  number={2},
  pages={269--287},
  year={1984},
  publisher={Cambridge University Press}
}

@article{cosmides1997modular,
  title={The modular nature of human intelligence},
  author={Cosmides, Leda and Tooby, John},
  journal={The origin and evolution of intelligence},
  pages={71--101},
  year={1997}
}

@article{keysers2019measuring,
  title={Measuring compositional generalization: A comprehensive method on realistic data},
  author={Keysers, Daniel and Sch{\"a}rli, Nathanael and Scales, Nathan and Buisman, Hylke and Furrer, Daniel and Kashubin, Sergii and Momchev, Nikola and Sinopalnikov, Danila and Stafiniak, Lukasz and Tihon, Tibor and others},
  journal={arXiv preprint arXiv:1912.09713},
  year={2019}
}

@article{wiedemer2023compositional,
  title={Compositional generalization from first principles},
  author={Wiedemer, Thadd{\"a}us and Mayilvahanan, Prasanna and Bethge, Matthias and Brendel, Wieland},
  journal={Advances in Neural Information Processing Systems},
  volume={36},
  pages={6941--6960},
  year={2023}
}

@article{zhao2023survey,
  title={A survey of large language models},
  author={Zhao, Wayne Xin and Zhou, Kun and Li, Junyi and Tang, Tianyi and Wang, Xiaolei and Hou, Yupeng and Min, Yingqian and Zhang, Beichen and Zhang, Junjie and Dong, Zican and others},
  journal={arXiv preprint arXiv:2303.18223},
  volume={1},
  number={2},
  year={2023}
}

@article{marcus2018deep,
  title={Deep learning: A critical appraisal},
  author={Marcus, Gary},
  journal={arXiv preprint arXiv:1801.00631},
  year={2018}
}

@article{patil2025advancing,
  title={Advancing reasoning in large language models: Promising methods and approaches},
  author={Patil, Avinash and Jadon, Aryan},
  journal={arXiv preprint arXiv:2502.03671},
  year={2025}
}

@article{cao2025large,
  title={Large language models for planning: A comprehensive and systematic survey},
  author={Cao, Pengfei and Men, Tianyi and Liu, Wencan and Zhang, Jingwen and Li, Xuzhao and Lin, Xixun and Sui, Dianbo and Cao, Yanan and Liu, Kang and Zhao, Jun},
  journal={arXiv preprint arXiv:2505.19683},
  year={2025}
}

@article{wang2024towards,
  title={Towards data-and knowledge-driven AI: a survey on neuro-symbolic computing},
  author={Wang, Wenguan and Yang, Yi and Wu, Fei},
  journal={IEEE Transactions on Pattern Analysis and Machine Intelligence},
  year={2024},
  publisher={IEEE}
}

@inproceedings{Lake2018,
  author    = {Lake, Brenden M. and Baroni, Marco},
  title     = {Generalization without Systematicity: On the Compositional Skills of Sequence-to-Sequence Recurrent Networks},
  booktitle = {International Conference on Machine Learning (ICML)},
  pages     = {2873--2882},
  year      = {2018}
}

@inproceedings{Kim2020,
  author    = {Kim, Najoung and Linzen, Tal},
  title     = {{COGS}: A Compositional Generalization Challenge Based on Semantic Interpretation},
  booktitle = {Proceedings of the 2020 Conference on Empirical Methods in Natural Language Processing (EMNLP)},
  pages     = {9107--9123},
  year      = {2020}
}

@inproceedings{Johnson2017,
  author    = {Johnson, Justin and Hariharan, Bharath and van der Maaten, Laurens and Fei-Fei, Li and Zitnick, C. Lawrence and Girshick, Ross},
  title     = {{CLEVR}: A Diagnostic Dataset for Compositional Language and Elementary Visual Reasoning},
  booktitle = {Proceedings of the {IEEE} Conference on Computer Vision and Pattern Recognition (CVPR)},
  pages     = {2901--2910},
  year      = {2017}
}

@article{Silver2016,
  author    = {Silver, David and Huang, Aja and Maddison, Chris J. and Guez, Arthur and Sifre, Laurent and van den Driessche, George and Schrittwieser, Julian and Antonoglou, Ioannis and Panneershelvam, Veda and Lanctot, Marc and Dieleman, Sander and Grewe, Dominik and Nham, John and Kalchbrenner, Nal and Sutskever, Ilya and Lillicrap, Timothy and Leach, Madeleine and Kavukcuoglu, Koray and Graepel, Thore and Hassabis, Demis},
  title     = {Mastering the Game of Go with Deep Neural Networks and Tree Search},
  journal   = {Nature},
  volume    = {529},
  number    = {7587},
  pages     = {484--489},
  year      = {2016}
}

@book{Russell2010,
  author    = {Russell, Stuart J. and Norvig, Peter},
  title     = {Artificial Intelligence: A Modern Approach},
  edition   = {3rd},
  publisher = {Pearson Education},
  year      = {2010}
}

@article{liang2025ai,
  title={AI Reasoning in Deep Learning Era: From Symbolic AI to Neural--Symbolic AI},
  author={Liang, Baoyu and Wang, Yuchen and Tong, Chao},
  journal={Mathematics},
  volume={13},
  number={11},
  pages={1707},
  year={2025},
  publisher={MDPI}
}

@inproceedings{mitra2015learning,
  title={Learning to automatically solve logic grid puzzles},
  author={Mitra, Arindam and Baral, Chitta},
  booktitle={Proceedings of the 2015 Conference on Empirical Methods in Natural Language Processing},
  pages={1023--1033},
  year={2015}
}

@inproceedings{piette2019ludii,
  title={Ludii and XCSP: playing and solving logic puzzles},
  author={Piette, C{\'e}dric and Piette, Eric and Stephenson, Matthew and Soemers, Dennis JNJ and Browne, Cameron},
  booktitle={2019 IEEE Conference on Games (CoG)},
  pages={1--4},
  year={2019},
  organization={IEEE}
}

@article{giadikiaroglou2024puzzle,
  title={Puzzle solving using reasoning of large language models: A survey},
  author={Giadikiaroglou, Panagiotis and Lymperaiou, Maria and Filandrianos, Giorgos and Stamou, Giorgos},
  journal={arXiv preprint arXiv:2402.11291},
  year={2024}
}

@article{berthier2013pattern,
  title={Pattern-based constraint satisfaction and logic puzzles},
  author={Berthier, Denis},
  journal={arXiv preprint arXiv:1304.1628},
  year={2013}
}

@inproceedings{qefalija2024literature,
  title={Literature Review on Constraint Satisfaction Problems Solving},
  author={Qefalija, Edlira and Snopce, Halil and Dermaku, Artan},
  booktitle={2024 8th International Symposium on Multidisciplinary Studies and Innovative Technologies (ISMSIT)},
  pages={1--6},
  year={2024},
  organization={IEEE}
}

@article{garcez2023neurosymbolic,
  title={Neurosymbolic ai: The 3 rd wave},
  author={Garcez, Artur d’Avila and Lamb, Luis C},
  journal={Artificial Intelligence Review},
  volume={56},
  number={11},
  pages={12387--12406},
  year={2023},
  publisher={Springer}
}

@article{sakr2022neuro,
  title={Neuro-symbolic AI star: A tale of two worlds},
  author={Sakr, Charbel and Hitzler, Pascal and Sheth, Amit},
  journal={AI Magazine},
  volume={43},
  number={4},
  pages={406--419},
  year={2022}
}

@article{hitzler2022neuro,
  title={Neuro-symbolic artificial intelligence: The state of the art},
  author={Hitzler, Pascal and Sarker, Md Kamruzzaman},
  year={2022},
  publisher={IOS press}
}

@article{Fodor1988,
  author    = {Fodor, Jerry A. and Pylyshyn, Zenon W.},
  title     = {Connectionism and Cognitive Architecture: A Critical Analysis},
  journal   = {Cognition},
  volume    = {28},
  number    = {1-2},
  pages     = {3--71},
  year      = {1988}
}

@inproceedings{Ruis2020,
  author    = {Ruis, Lluis and Andreas, Jacob and Baroni, Marco and Dagan, Idan and Goldberg, Yoav},
  title     = {A Benchmark for Systematic Generalization in Grounded Language Understanding},
  booktitle = {International Conference on Machine Learning (ICML)},
  year      = {2020}
}

@inproceedings{Keysers2020,
  author    = {Keysers, Daniel and Schärli, Nathanael and Kale, Nitish and Cer, Daniel and Firat, Orhan and Mourad, Ayoub and Riesa, Jason and Bapna, Ankur and Caswell, Isaac and Hassan, Hany},
  title     = {Measuring Compositional Generalization: A Comprehensive Method on Realistic Data},
  booktitle = {International Conference on Learning Representations (ICLR)},
  year      = {2020}
}

@inproceedings{Yi2020,
  author    = {Yi, Kexin and Gan, Chuang and Li, Yilun and Torralba, Antonio and Kohli, Pushmeet and Tenenbaum, Josh},
  title     = {{CLEVRER}: CoLlision Events for Video REpresentation and Reasoning},
  booktitle = {International Conference on Learning Representations (ICLR)},
  year      = {2020}
}

@inproceedings{Hudson2019,
  author    = {Hudson, Drew A. and Manning, Christopher D.},
  title     = {{GQA}: A New Dataset for Real-World Visual Reasoning and Compositional Question Answering},
  booktitle = {Proceedings of the {IEEE/CVF} Conference on Computer Vision and Pattern Recognition (CVPR)},
  year      = {2019}
}

@article{Brown2019,
  author    = {Brown, Noam and Sandholm, Tuomas},
  title     = {Superhuman {AI} for Multiplayer Poker},
  journal   = {Science},
  volume    = {365},
  number    = {6456},
  pages     = {885--890},
  year      = {2019}
}

@inproceedings{Manhaeve2018,
  author    = {Manhaeve, Robin and Dumančić, Sebastijan and Kimmig, Angelika and Demeester, Thomas and De Raedt, Luc},
  title     = {{DeepProblog}: Neural Probabilistic Logic Programming},
  booktitle = {Advances in Neural Information Processing Systems (NeurIPS)},
  year      = {2018}
}

@inproceedings{Dong2019,
  author    = {Dong, Honghua and Mao, Jiayuan and Lin, Tian and Wang, Chong and Lih, Lihong and Zhou, Denny},
  title     = {Neural Logic Machines},
  booktitle = {International Conference on Learning Representations (ICLR)},
  year      = {2019}
}

@inproceedings{Wang2019,
  author    = {Wang, Po-Wei and Donti, Priya and Wilder, Bryan and Kolter, Zico},
  title     = {{SATNet}: Bridging Deep Learning and Logical Reasoning Using a Differentiable {MAX-SAT} Solver},
  booktitle = {International Conference on Machine Learning (ICML)},
  year      = {2019}
}

@misc{Riegel2020,
  author    = {Riegel, Ryan and Gray, Alexander and Luus, Francois and Khan, Naila and Makondo, Nolo and Akhalwaya, Ismail Yunus and Yarkoni, T. and Koco, Z. and Mvelase, P.},
  title     = {Logical Neural Networks},
  year      = {2020},
  eprint    = {2006.13155},
  archivePrefix = {arXiv}
}

@inproceedings{Andreas2016,
  author    = {Andreas, Jacob and Rohrbach, Marcus and Darrell, Trevor and Klein, Dan},
  title     = {Neural Module Networks},
  booktitle = {Proceedings of the {IEEE} Conference on Computer Vision and Pattern Recognition (CVPR)},
  year      = {2016}
}

\end{document}